\documentclass[10pt, twocolumn]{article}
\usepackage[T1]{fontenc}
\usepackage{graphicx}
\usepackage{amsmath}  
\usepackage{amssymb}  
\usepackage{amsfonts} 
\usepackage{hyperref} 
\usepackage{glossaries}
\usepackage{float}
\usepackage[textsize=small]{todonotes}
\usepackage{xcolor}

\newcommand{\bbN}{\mathbb{N}}
\newcommand{\bbR}{\mathbb{R}}

\newcommand{\bbRm}{\mathbb{R}^m}
\newcommand{\bbRk}{\mathbb{R}^k}

\newcommand{\bbRkm}{\mathbb{R}^{k \times m}}
\newcommand{\mcX}{\mathcal{X}}
\newcommand{\mcY}{\mathcal{Y}}
\newcommand{\mcC}{\mathcal{C}}
\newcommand{\mcZ}{\mathcal{Z}}
\newcommand{\mcZp}{\mathcal{Z}^+}
\newcommand{\mcZn}{\mathcal{Z}^-}
\newcommand{\mbx}{\mathbf{x}}

\newcommand{\mbz}{\mathbf{z}}
\newcommand{\mbt}{\mathbf{t}}
\newcommand{\mbw}{\mathbf{w}}
\newcommand{\mbb}{\mathbf{b}}
\newcommand{\mbc}{\mathbf{c}}
\newcommand{\lcav}{\mathcal{L}_{\text{CAV}}}
\newcommand{\lorth}{\mathcal{L}_{\text{orth}}}
\newcommand{\lalphabeta}{\mathcal{L}_{\text{orth}}^{\beta}}

\usepackage{xr}

\makeatletter

\usepackage{xspace}

\DeclareRobustCommand\onedot{\futurelet\@let@token\@onedot}
\def\@onedot{\ifx\@let@token.\else.\null\fi\xspace}

\def\eg{\emph{e.g}\onedot} 
\def\ie{\emph{i.e}\onedot}

\newacronym{cav}{\mbox{CAV}}{Concept Activation Vector}
\newacronym{svm}{\mbox{SVM}}{Support Vector Machine}
\newacronym{auroc}{\mbox{AUROC}}{Area Under Receiver Operating Curve}
\newacronym{lrp}{\mbox{LRP}}{Layer-wise Relevance Propagation}
\newacronym{clarc}{\mbox{ClArC}}{Class Artifact Compensation}
\newacronym{pclarc}{\mbox{P-ClArC}}{Projective Class Artifact Compensation}
\newacronym{xai}{\mbox{XAI}}{eXplainable Artificial Intelligence}
\newacronym{dnn}{\mbox{DNN}}{Deep Neural Network}
\newacronym{pca}{\mbox{PCA}}{Principal Component Analysis}
\newacronym{ica}{\mbox{ICA}}{Independent Component Analysis}

\usepackage{hyperref}
\glsdisablehyper

\title{
\vspace{-2em}\textbf{Post-Hoc Concept Disentanglement: From Correlated to Isolated Concept Representations}}

\author{
    Eren Erogullari\textsuperscript{\rm 1},
    Sebastian Lapuschkin\textsuperscript{\rm 1,2,$\dagger$},
    Wojciech Samek\textsuperscript{\rm 1,3,4,$\dagger$},
    Frederik Pahde\textsuperscript{\rm 1,$\dagger$}
}
\date{\small
    \textsuperscript{\rm 1}Fraunhofer Heinrich Hertz Institut, Berlin, Germany\\
    \textsuperscript{\rm 2}Centre of eXplainable Artificial Intelligence, Technological University Dublin \\
    \textsuperscript{\rm 3}Technische Universität Berlin, Berlin, Germany\\
    \textsuperscript{\rm 4}Berlin Institute for the Foundations of Learning and Data (BIFOLD), Berlin, Germany\\
    \textsuperscript{\rm $\dagger$}corresponding authors: \texttt{\{sebastian.lapuschkin,wojciech.samek,frederik.pahde\}@hhi.fraunhofer.de}\\
}

\begin{document}
\maketitle              

\setcounter{footnote}{0} 
\begin{abstract}
Concept Activation Vectors (CAVs) are widely used to model human-understandable concepts as directions within the latent space of neural networks. They are trained by identifying directions from the activations of concept samples to those of non-concept samples. However, this method often produces similar, non-orthogonal directions for correlated concepts, such as ``beard'' and ``necktie'' within the CelebA dataset, which frequently co-occur in images of men. This entanglement complicates the interpretation of concepts in isolation and can lead to undesired effects in CAV applications, such as activation steering.
To address this issue, we introduce a post-hoc concept disentanglement method that employs a non-orthogonality loss, facilitating the identification of orthogonal concept directions while preserving directional correctness. We evaluate our approach with real-world and controlled correlated concepts in CelebA and a synthetic FunnyBirds dataset with VGG16 and ResNet18 architectures. We further demonstrate the superiority of orthogonalized concept representations in activation steering tasks, allowing (1) the \emph{insertion} of isolated concepts into input images through generative models and (2) the  \emph{removal} of concepts for effective shortcut suppression with reduced impact on correlated concepts in comparison to baseline CAVs.\footnote{Code is available at \url{https://github.com/erenerogullari/cav-disentanglement}}
\end{abstract}

\section{Introduction}
\label{introduction}
With the growing reliance on deep learning in critical domains, such as applications in medicine~\cite{brinker2019deep} or criminal justice~\cite{zavrvsnik2021algorithmic,travaini2022machine}, the need for \gls{xai} has become more relevant to ensure transparency and trust in model's decision-making processes. 
\gls{xai} methods can be broadly categorized into local approaches, explaining individual predictions by attributing importance scores to input features (\eg,~\cite{bach2015pixel,selvaraju2017grad,lundberg2017unified}), and global approaches, aiming to uncover broader decision-making patterns learned by a model (\eg,~\cite{kim2018interpretability,achtibat2023attribution,fel2023craft}).
As such, concept-based explanations seek to represent abstract, human-understandable concepts in the model's latent space, offering insights into how these concepts influence predictions. 
Specifically,  \glspl{cav} allow interpreting \glspl{dnn} by modeling high-level concepts as directions in latent space \cite{kim2018interpretability}. 
They are typically learned using linear classifiers, \eg, linear \glspl{svm}, to separate activations corresponding to the presence vs absence of a concept, represented by the normal to the decision hyperplane of the classifier.
However, while this approach optimizes the directional correctness of \emph{individual} concepts, it may fail to isolate a concept’s direction from others when multiple correlated concepts are trained simultaneously.
This is caused by highly entangled neural representations   \cite{elhage2022toy}, \ie, multiple concepts being encoded along overlapping directions due to correlations present in the dataset. 
\begin{figure*}[t]
    \centering
    \includegraphics[width=.8\textwidth]{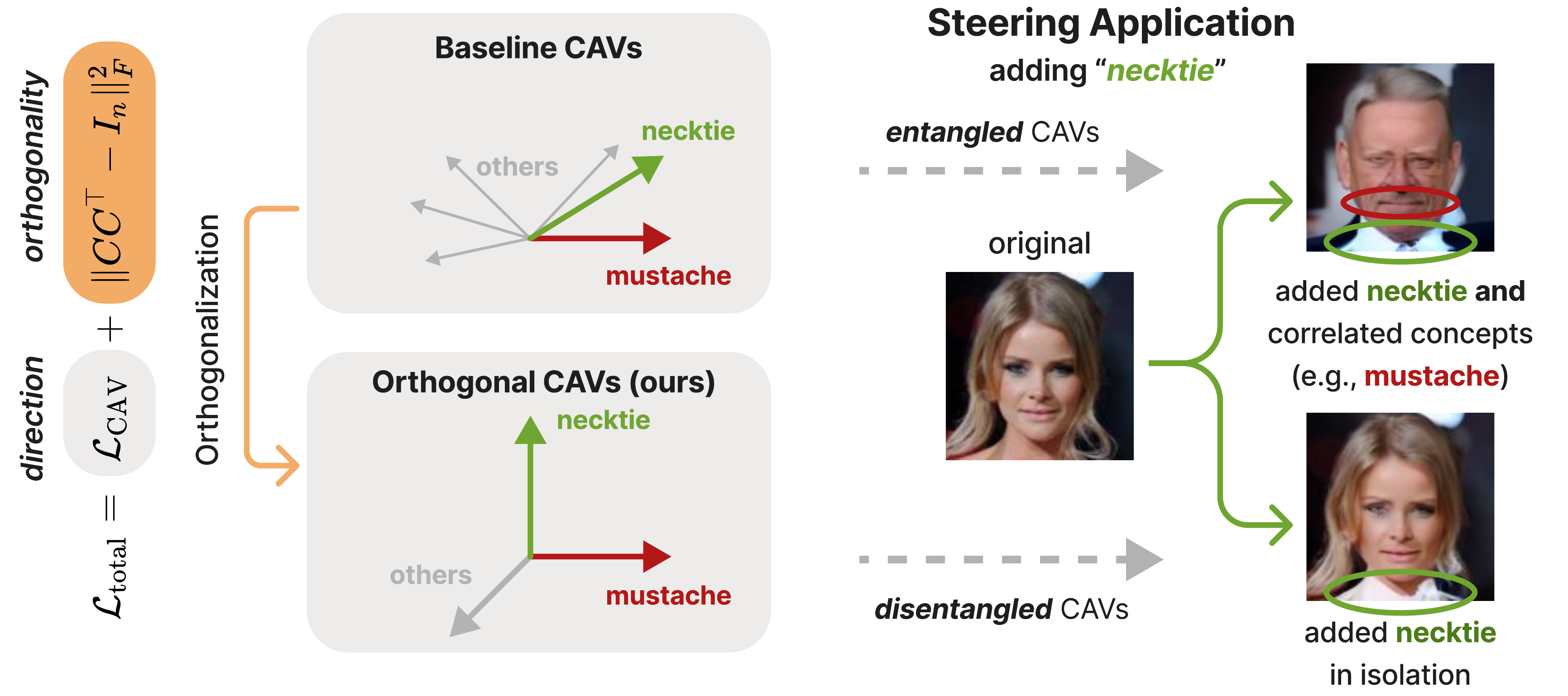}
    \caption{
    \emph{Left:} Our novel \gls{cav} objective encourages the orthogonalization of multiple concept directions trained simultaneously. 
    \emph{Right:} The resulting disentangled \glspl{cav} are beneficial for various \gls{cav} applications, as concepts can be targeted in isolation. 
    For example, when inserting the ``necktie'' concept to an input image in a steering task, the usage of \emph{entangled} baseline \glspl{cav} might add correlated concepts as well (\eg, ``mustache''), while \emph{disentangled} \glspl{cav} add the targeted concept in isolation.
    }
    \label{fig:main_figure}
\end{figure*}
This entanglement manifests in latent space as non-orthogonal concept directions~\cite{nicolson2025explaining}, making it challenging to isolate individual concepts and leading to ambiguity and less interpretability in \gls{cav}-based explanations.
For example, \gls{cav}-based steering applications, attempting to add~\cite{preechakul2022diffusion} or remove~\cite{anders2022finding} specific concepts, may unintentionally modify other entangled concepts, as illustrated in Fig.~\ref{fig:main_figure}~(\emph{top~right}).

To tackle this issue holistically, we propose a novel \gls{cav} training objective penalizing non-orthogonality between concept directions trained simultaneously, thereby encouraging disentangled representations in latent space (Fig.~\ref{fig:main_figure}, \emph{left}). 
Our proposed loss term can be utilized in conjunction with any objective targeting directional correctness in a weighted manner to balance the trade-off between both optimization goals.
We further introduce targeted orthogonalization, allowing to selectively enforce separation between specific concept pairs. 

We evaluate our post-hoc concept orthogonalization approach through both controlled and real-world experiments using the CelebA and FunnyBirds datasets with VGG16 and ResNet18 architectures. 
Furthermore, we demonstrate the effectiveness of reduced concept entanglement in activation steering applications. This includes (1) the insertion of isolated concepts in latent encodings of contemporary Diffusion models as shown in Fig.~\ref{fig:main_figure}~(\emph{bottom~right}) and (2) the precise concept removal for shortcut suppression with minimal impact on correlated concepts. 
We compare orthogonalized CAVs to baseline CAVs trained in isolation and provide both qualitative and quantitative results.

\section{Related Work}
\label{related-work}
Existing works either interpret concepts as individual neurons~\cite{achtibat2023attribution, olah2017feature}, higher-dimensional subspaces \cite{vielhaben2023multi}, or linear directions within the latent space~\cite{nanda2023emergent, kim2018interpretability, pahde2022navigating, brocki2019concept}.
The latter provides a flexible way to capture representations by interpreting them as superpositions of multiple neurons \cite{elhage2022toy}. 
Matrix factorization methods, such as non-negative matrix factorization, can be utilized to extract meaningful basis components in an unsupervised manner that act as interpretable concepts~\cite{fel2023craft, zhang2021invertible}. 
We adopt the linear-directions paradigm and interpret concepts as linear directions learned in a supervised manner from latent model activations. 
As many concept-based methods, such as steering methods for concept insertion or suppression, scale to multiple concepts, ensuring that learned directions remain disentangled becomes crucial to avoid interfering concepts.

Concept disentanglement methods can generally be categorized into (1) approaches utilized before or during model training, aiming to learn disentangled representations from the beginning, and (2) post-hoc disentanglement approaches, focusing on disentangling  already learned but entangled concepts. 
Pre- or during-training strategies introduce constraints like whitening layers to ensure concept disentanglement across model's layers \cite{chen2020concept}, leverage metadata to separate relevant features from biases \cite{rakowski2024metadata}, or focus on learning disentangled representations of the underlying concepts in the data \cite{wang2024disentangled}, either in a supervised \cite{bouchacourt2018multi, trauble2021disentangled} or in an unsupervised manner \cite{singh2019finegan, hjelm2019learning, kim2018disentangling}. 
Post-hoc methods, on the other hand, typically extend classic dimensionality-reduction techniques like \gls{pca} or \gls{ica} \cite{chormai2024disentangled} to uncover disentangled directions from pretrained models without altering their original training pipelines.
In contrast to other methods, we introduce a \emph{supervised post-hoc} concept disentanglement approach by extending \glspl{cav} \cite{kim2018interpretability, pahde2022navigating} to enforce orthogonal concept directions within the latent space. 

\section{Post-Hoc Concept Orthogonalization}
\label{methods}
\paragraph{Notation.} 
Given a neural network $f: \mcX \to \mcY$ that maps input samples $\mbx \in \mcX$ to target labels $y \in \mcY$, we can decompose the network into two functions $f = h \, \circ \, g$, with feature extractor $g: \mcX \to \mcZ$ with $\mcZ \subseteq \bbRm$ mapping input samples to hidden layer activations $\mbz \in \mcZ$ at a given layer with $m$ neurons, and classifier $h: \mcZ \to \mcY$ mapping hidden layer activations to target labels $y$. 
Furthermore, given $n \in \bbN$ concepts with  binary concept labels $\mbt^{(i)} \in \{ -1, 1\}^{n}$ for each sample $\mbx^{(i)} \in \mcX$, the set of latent activations $\mcZ$ of samples can be partitioned into two sets for each concept $c$ with $\mcZ=\mcZp_c \cup \mcZn_c$, where $\mcZp_c := \{ g(\mbx^{(i)}) \, | \, \mbx^{(i)}\in \mcX \, \text{ and } t^{(i)}_c = 1\, \}$ is the set of activations with the target concept c, and $\mcZn_c := \{ g(\mbx^{(i)}) \, | \, \mbx^{(i)}\in \mcX \, \text{ and } t^{(i)}_c = -1\, \}$ is the set of activations without the target concept. 
Whereas the choice of layer is problem-specific, in practice, commonly the penultimate layer is used, as research suggests that later layers have higher receptive fields and capture more complex and abstract concepts \cite{bau2020understanding,olah2017feature,radford2017learning}.

\subsection{Concept Modeling via Concept Activation Vectors}
\glspl{cav} are defined as the direction in latent space pointing from activations of samples \emph{without} the target concept $\mcZn$ to activations of samples \emph{with} the target concept $\mcZp$~\cite{kim2018interpretability}.
Most commonly,  they are obtained from the weight vector $\mbw \in \bbRm$ of a linear classifier \cite{haufe2014interpretation,yuksekgonul2022post}, \ie, the hyperplane separating latent activations of these sample sets, some of which are linear \glspl{svm} minimizing the hinge loss with L2 regularization \cite{cortes1995support} and logistic, or ridge regression models minimizing the squared error loss with L2 regularization \cite{hoerl2000ridge}. 
By finding the optimal classification boundary, linear classifiers aim to achieve the directional correctness of concepts.

Given $k=|\mcX| \in \bbN$ samples, a concept $c$, and a simple linear model $f_{\text{lin}}(\mbz) = \mbw^T \mbz + b$ with weight vector $\mbw \in \bbRm$ and bias $b \in \bbR$ as the classifier along with a ridge regression term, we typically obtain the objective function
\begin{multline}
\label{eq:linear_loss}
    \arg\min_{\mbw, b} \, \mathcal{L} (Z ; \mbw, b) = \\
    \arg\min_{\mbw, b} \, \| \mbt - Z\mbw - \mbb \|_2^2 + \| \mbw \|_2^2
\end{multline}
with $\mbt \in \{ -1, 1\}^k$ representing the label vector with concept label $t_c^{(i)}$ as the $i^{th}$ element, $\mbb \in \bbRk$ as an $n$-wise repetition of $b$, and $Z \in \bbRkm$ as the input matrix with latent activations $\mbz^{(i)} \in \bbRm$ on its rows. 

Eq.~\eqref{eq:linear_loss} aims to find a weight that maximizes class-separability, which is prone to capturing distractor components that arise from noise and unrelated features in the data \cite{haufe2014interpretation}. 
To address this, Pahde et al.~\cite{pahde2022navigating} introduce Pattern-\glspl{cav}, assuming a linear dependency between the activations and the concept labels and aiming to find a pattern that explains $Z$ w.r.t concept labels $\mbt$, where the objective becomes 
\begin{multline}
\label{eq:signal-cav}
    \arg\min_{\mbw, \mbb} \, \lcav \, (Z \, ; \, \mbw, \mbb) = \\
    \arg\min_{\mbw, \mbb} \, || \, Z - \mbt\mbw^T -\mbb \, ||_2^2 
\end{multline}
leading to a solution invariant under feature scaling and more robust to noise. 

\begin{figure*}[t]
    \centering
    \includegraphics[width=.9\textwidth]{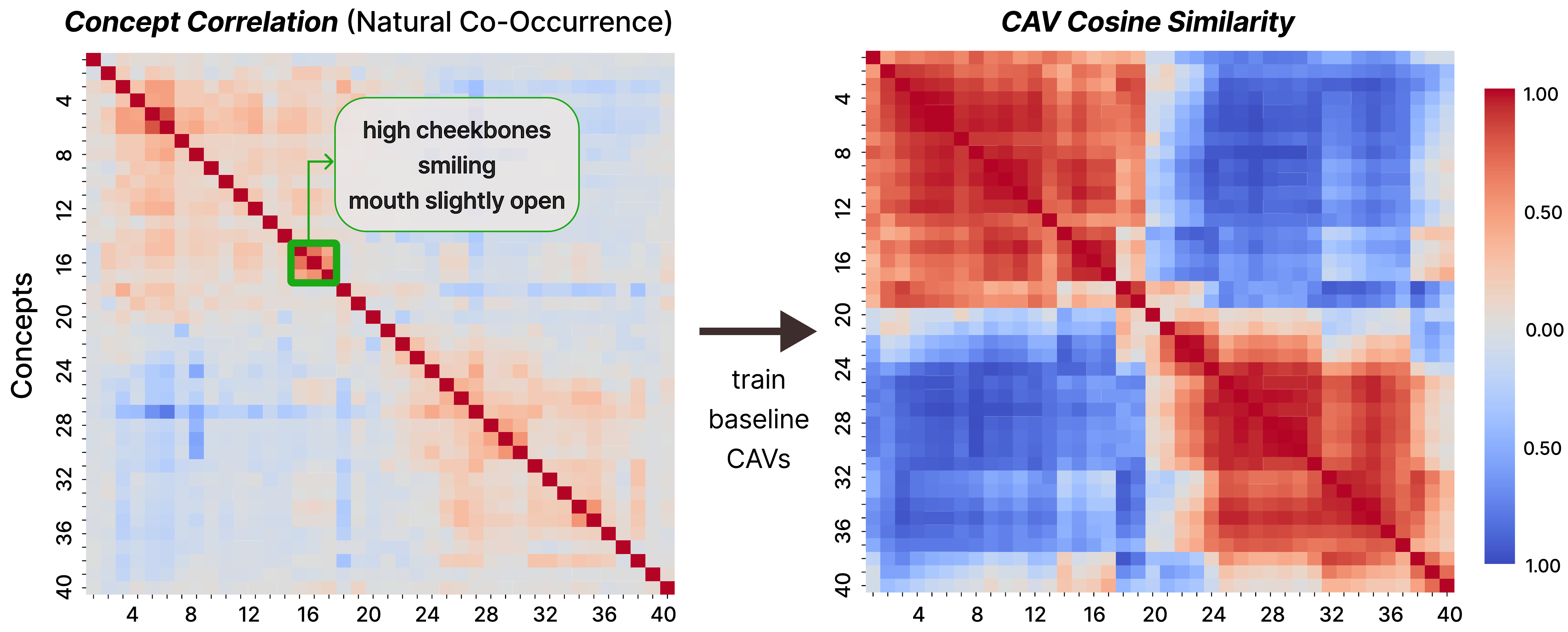}
    \caption{\emph{Left:} Correlations of known concepts based on their co-occurrence in CelebA. 
    \emph{Right:}
    Pair-wise cosine similarities between concept representations via \glspl{cav} trained in isolation. Concepts frequently co-occurring in the training data (\eg., ``high cheekbones'', ``smiling'', and ``mouth slightly open'') result in highly similar and entangled \glspl{cav}.
    }
    \label{fig:concept_correlation_and_entanglement}
\end{figure*}
However, both perspectives optimize \glspl{cav} in isolation and independent of other concepts, such that concept directions have no impact on other \glspl{cav}. 
This can result in solutions where \glspl{cav} share similar orientations and become entangled due to existing correlations in the training data.
Throughout this paper, we will utilize Pattern-CAVs, as defined in  Eq.~\eqref{eq:signal-cav}, as the \emph{baseline} \glspl{cav}. 

\subsection{Measuring Concept Entanglement}
We consider concepts as entangled if their \glspl{cav} share similar orientations within the latent space, which can be analyzed empirically by their cosine similarity matrix. 
The cosine similarity between two vectors measures their angular similarity and, given two \glspl{cav} $\mathbf{c}_i$ and $\mathbf{c}_j$, is defined as
\begin{equation}
\label{eq:cos-sim}
    \text{cos}(\mathbf{c}_i, \mathbf{c}_j) = \frac{\mathbf{c}_i \cdot \mathbf{c}_j}{\|\mathbf{c}_i\| \, \|\mathbf{c}_j\|}
\end{equation}
where values near 1 indicate high alignment, \ie, entanglement, between the concepts, values near -1 indicate a strong inverse correlation or opposing directions in the feature space, and values near 0 suggest orthogonality and independence. 
The resulting matrix $C = (C_{ij}) =  (\text{cos}(\mathbf{c}_i, \mathbf{c}_j))$ provides insight into concept entanglement, where highly similar concepts may occupy overlapping regions in latent space. To quantify the orthogonality of a given concept to all others, we use the average cosine similarity, effectively describing the orthogonality of a concept in terms of a single metric, defined in Eq.~\eqref{eq:uniqueness}. 

An example for highly entangled concepts is shown in Fig.~\ref{fig:concept_correlation_and_entanglement}, where we show the \glspl{cav} trained on the CelebA dataset~\cite{liu2015faceattributes}, consisting of images of celebrity faces with concept level annotations, such as ``beard'', ``mustache'', and ``makeup''. 
The resulting cosine similarity matrix of \glspl{cav} contains naturally emerging entanglement blocks, where different sets of concepts point in similar or opposite directions. 
This entanglement typically arises from the training of individual concepts in isolation without any information on other concepts, such that correlations in the dataset, \eg, concepts that frequently appear together (\eg, ``beard'' and ``mustache'') or those that do not (\eg, ```mustache'' and ``makeup''), cause \glspl{cav} to align in similar or opposite directions in the latent space. 

\paragraph{Orthogonality Metric.}
Given a finite set $\mcC$ of concepts, we define the \textit{orthogonality} $O_i$ of a concept $ \mathbf{c}_i \in \mcC$ as:

\begin{equation}
\label{eq:uniqueness}
O_{i} = 1 - \frac{1}{|\mcC|-1} \sum_{\tilde{\mbc} \in \mcC \setminus \{\mbc_i\}} |\cos(\mbc_i, \tilde{\mbc})|
\end{equation}

where $|\cdot|$ is the absolute value function. Consequently, the value of $O_i$ ranges between 0 and 1, where 1 represents perfect disentanglement and orthogonality of the given concept, and 0 represents complete entanglement and alignment with other concepts, sharing the same orientation in the latent space. We further define \textit{average orthogonality} $\bar{O}$ as the average of all concepts' orthogonality values and use it as an indicator of overall disentanglement of concepts.

\subsection{Orthogonalization of \glspl{cav}}
\label{non_orthogonality_penalization}
Given a hidden layer with $m \in \bbN$ neurons and a finite set of concepts $\mcC$ present in the dataset with \mbox{$n = |\mcC|$} concepts, and further assuming a setting where the dimensionality of the latent space is significantly higher than the number of concepts, \ie, when $m \gg n$, there is sufficient room for \glspl{cav} to be orthogonal to one another.
Therefore, to disentangle concept representations, we propose an 
additional loss term, encouraging orthogonality between \glspl{cav} in latent space. Specifically, we define the orthogonality loss $\lorth$ as

\begin{equation}
\label{eq:orth_cav}
    \lorth = \| C C^\top - I_n \|_F^2 
\end{equation}

where $C \in \mathbb{R}^{n \times m}$ is the matrix of $n$ \glspl{cav} with each row corresponding to a \gls{cav}, $I_n$ is the $n$-dimensional identity matrix and $\| \cdot \|_F^2$ denotes the squared Frobenius norm.
This loss term penalizes the deviation of the pair-wise cosine similarity matrix $C C^\top$ from the identity matrix $I_n$, effectively encouraging the \glspl{cav} to be orthogonal.
The orthogonality loss $\lorth$ can be combined with the original loss term $\lcav$ (e.g., in Eq.~\eqref{eq:signal-cav}) in a weighted manner as:

\begin{equation}
\label{eq:total_loss}
    \mathcal{L} = \lcav + \alpha \, \lorth
\end{equation}

yielding the minimization objective with weighting parameter $\alpha > 0$ balancing the trade-off between the potentially competing goals to (1) maximize directional correctness with $\lcav$ and (2) orthogonalize the \glspl{cav} with $\lorth$. 
With $\alpha = 0$ the optimization objective becomes the original objective, which does not penalize the concept entanglement, whereas with $\alpha \rightarrow \infty$ the orthogonality term will dominate the loss and the optimization will most likely yield random orthogonal concept directions that fail at directional correctness. 

In practice, our novel training objective can either be employed to optimize \glspl{cav} from random initialization or to disentangle pre-trained \glspl{cav} in a fine-tuning step.
While the former can balance the directional correctness objective and the orthogonality constraint from the beginning, it may require more iterations for convergence. 
The latter approach can lead to faster convergence, however, the magnitude of $\alpha$ must be chosen carefully to prevent over-correction by the orthogonality term.
Finally, the average orthogonality $\bar{O}$ can be used as a global metric to measure overall orthogonality of \glspl{cav} during optimization.

\subsection{Targeted Orthogonalization with Weighted Penalization}
\label{alpha_beta_method}
While orthogonalization of all \glspl{cav} is beneficial, not all concepts require the same level of adjustment. For example, already disentangled concepts, \ie, minimally correlated concepts, should remain largely unaffected, while entangled concepts should be prioritized. 
To achieve this, we introduce a symmetric weighting matrix $W_{\beta} \in \mathbb{R}^{n \times n}$ to adjust the weighting of concept pairs given a set of target pairs $\mathcal{T}$, resulting in the $\beta$-weighted orthogonalization loss $\lalphabeta$ defined as:

\begin{equation}
\label{eq:alpha_beta_loss}
    \lalphabeta = \| W_{\beta} \odot (C C^\top - I_n) \|_F^2
\end{equation}
where
\begin{multline}
\label{eq:w_alpha_beta}
(W_{\beta})_{ij} = (W_{\beta})_{ji} =
        \begin{cases} 
        \beta & , \text{if }(i, j) \in \mathcal{T} \\
        1 & ,\text{otherwise}.
        \end{cases}
\end{multline}
where $\beta \in \bbR$ with $\beta>0$ describing the relative importance of target pairs over non-target pairs and $\odot$ denoting the Hadamard (element-wise) product. Having $\beta > 1$ will enforce a stricter disentanglement on the target pairs, while having $0<\beta < 1$ will effectively relax the orthogonality constraint. 
Plugging the new $\lalphabeta$ loss in to Eq.~\eqref{eq:total_loss} yields a new optimization objective, penalizing non-orthogonality of selected pairs of concepts over the others.
This formulation allows selective penalization, focusing on disentangling specific pairs of highly entangled concepts while leaving already disentangled pairs largely unaffected. 
Note, that this formulation can easily be extended to individual weights for each pair, \ie different values for each entry in $W_{\beta}$, while keeping $W_{\beta}$ symmetric. 
In practice, target pairs can be defined as a list of most entangled pairs with high correlations or high cosine similarities.

\subsection{Practical Implications: Directional Correctness and Orthogonality Trade-off}
In order to maximize concept orthogonality while maintaining directional correctness, we monitor the macro-averaged \gls{auroc}\footnote{Note, that although our \glspl{cav} are \emph{not} computed as predictors, their dot products with latent activations can be used to measure the concept separability.}, as a proxy metric for directional correctness, together with average orthogonality. 
 Monitoring \gls{auroc} allows to ensure that the optimization does not compromise the primary goal of \glspl{cav}.
 We can further use \gls{auroc} to define an early-exit criteria, \eg, thresholds based on the average AUROC, average-drop of AUROC, or max-drop of AUROC.
 This ensures that the concept disentanglement preserves the directional correctness by preventing over-optimization at the expense of predictive performance, ensuring the CAVs remain useful for their intended purpose.
 \begin{figure*}[h]
    \centering
    \includegraphics[width=\textwidth]{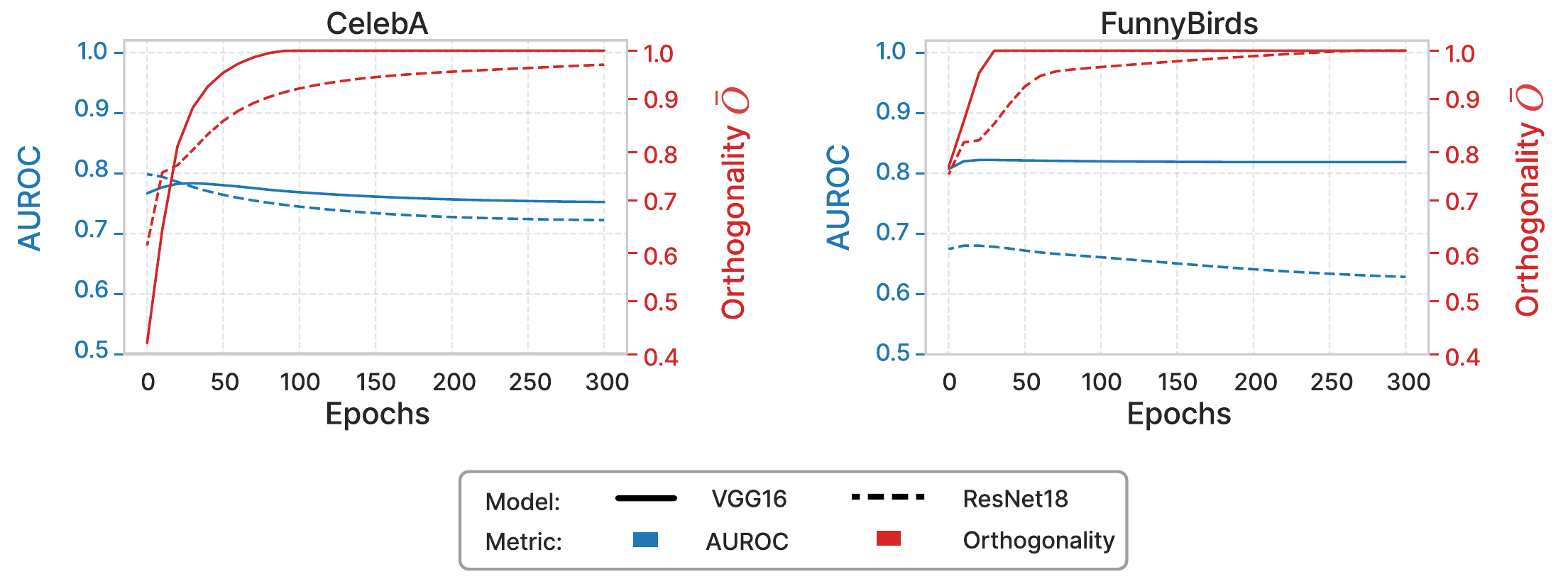}
    \caption{Evolution of AUROC (\emph{blue}) and average orthogonality $\bar{O}$ (\emph{red}) during \gls{cav} optimization for ResNet18 and VGG16 models trained on CelebA (\emph{left}) and FunnyBirds (\emph{right}). Our approach achieves near-perfect orthogonalization, while preserving directional correctness as measured via AUROC.}
    \label{fig:all_models_metrics}
\end{figure*}

\section{Experiments}
\label{experiments}
We empirically evaluate our post-hoc concept disentanglement methods using real-world concepts in CelebA and controlled concepts in the synthetic \mbox{FunnyBirds} dataset. 
Specifically, our experiments investigate whether our approach can successfully disentangle \gls{cav}-based concept representations while preserving the directional correctness measured via their \gls{auroc} as a proxy metric.

\subsection{Experiment Details}
\label{experiment-details}
We conduct experiments with the real-world CelebA dataset~\cite{liu2015faceattributes} and the synthetic FunnyBirds dataset~\cite{hesse2023funnybirds}. 
The former consists of images of faces along with binary concept labels for 40 attributes (see Appendix~\ref{appendix:celeba_details} for details).
We consider the task of classifying samples with and without ``blond hair''.
Fig.~\ref{fig:concept_correlation_and_entanglement} (\emph{left}) presents the correlation matrix between these attributes, revealing two distinct blocks that reflect natural groupings of concepts associated with male and female attributes. 
Moreover, we utilize a synthetic dataset with generated images of 50 bird classes with part-level annotations for 5 body parts (beak, eye, foot, tail, and wing), which are transformed into binary concept labels (see Appendix~\ref{appendix:funnybirds_details} for details). 
We inject controlled correlations between ``beak'' and ``tail'' concepts, such that $70\%$ of samples with a given beak type have a specific corresponding tail type. 
For example, beak type ``beak01.glb'' has a $70\%$ probability of being paired with tail type ``tail01.glb'', while the remaining $30\%$ is split equally between the other tail types. This pattern is similarly applied to other beak types, resulting in controlled correlations between these concept attributes. 
For both datasets, we train VGG16 \cite{simonyan2014very} and ResNet18~\cite{he2016deep} models and train \glspl{cav} using activations from the last convolutional layers of both models. 

\subsection{Concept Disentanglement}
To measure the orthogonality of \glspl{cav} and directional correctness during the proposed optimization procedure, we (1) monitor the per-concept orthogonality $O_i$, average orthogonality $\bar{O}$, as well as per-concept and macro-averaged \gls{auroc} during optimization and (2) compare the cosine similarity matrices between \glspl{cav} before and after optimization. 
In both experimental settings we optimize \glspl{cav} starting from both random and pre-trained initializations, \ie, the solution for Eq.~\eqref{eq:signal-cav}, and apply orthogonalization, as defined in Eq.~\eqref{eq:total_loss}.

\paragraph{Orthogonalization and Directional Correctness.}
\label{orthogonalization_directional_correctness}
First, we evaluate our proposed optimization objective on all datasets using both models. 
We fine-tune pre-trained \glspl{cav} for $300$ epochs (see Appendix~\ref{appendix:cav_training_details} for details). 
The results are shown in Fig.~\ref{fig:all_models_metrics}.
Across all experiments, we observe a drastic increase in orthogonality, with \glspl{cav} achieving either near-perfect or complete orthogonalization within $300$ epochs. This strong shift towards orthogonality indicates that our method effectively promotes concept disentanglement, ensuring that learned representations become more independent. 

We further observe that, while enforcing orthogonality, directional correctness -- measured via \mbox{AUROC} as a proxy metric -- remains largely preserved. 
This indicates that the orthogonalization of \glspl{cav} only minimally harms their primary objective.  
Notably, in most experiments AUROC exhibits an initial increase at the beginning of optimization before undergoing a gradual decay and finally stabilizing at a slightly lower level. The initial increase in AUROC is likely due to the orthogonality constraint, which provides the optimization with a term that helps \glspl{cav} to be optimized jointly, as opposed to the baseline objective where each \gls{cav} is optimized separately. This encourages a better use of latent space via a more structured adjustment of \glspl{cav}. 
Overall, our results demonstrate that concept disentanglement can be effectively achieved without compromising directional correctness, making this approach robust across various settings.
\begin{figure*}[t]
    \centering
    \includegraphics[width=1\textwidth]{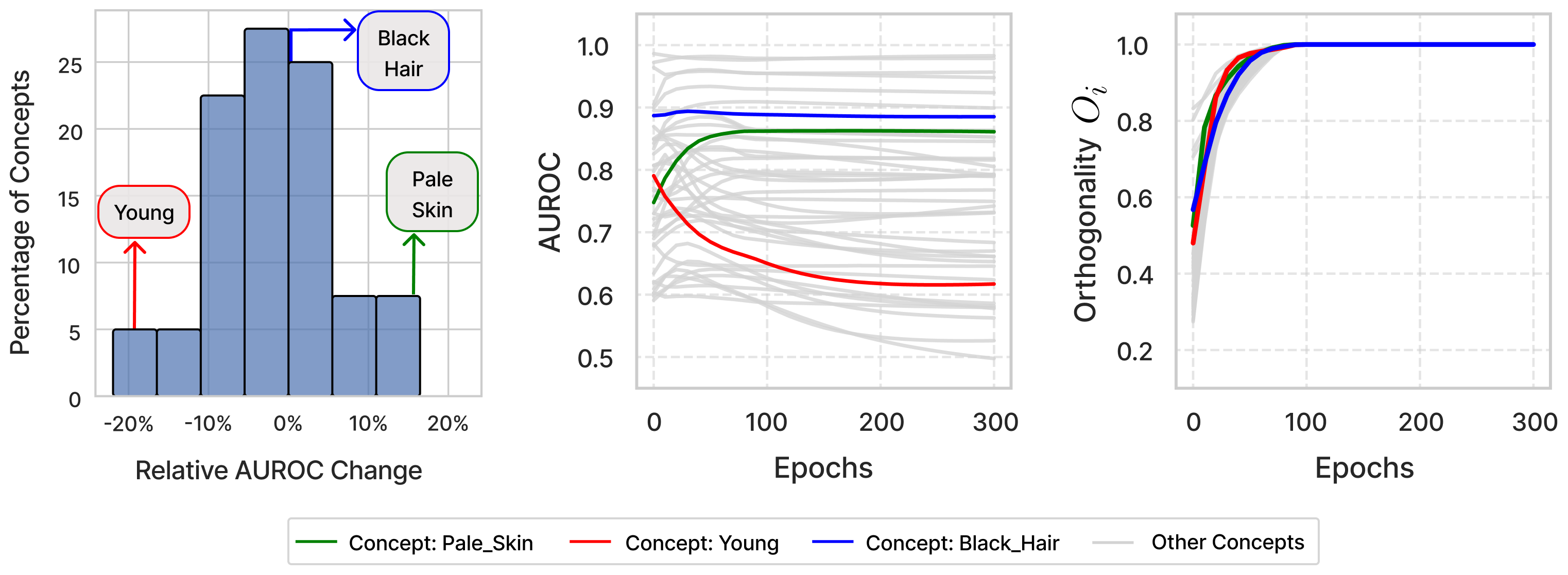}
    \caption{ \emph{Left}: Distribution of relative \gls{auroc} change before and after \gls{cav} optimization for the VGG16 model for CelebA. \emph{Middle and Right}: Evolution of per-concept metrics orthogonality $O_i$ and \gls{auroc}. 
    We highlight concepts with the highest increase (\emph{green}), the highest decrease (\emph{red}), and the smallest change in \gls{auroc} (\emph{blue}).}
    \label{fig:individual_metrics}
\end{figure*}

\paragraph{Concept Dynamics Under Orthogonalization.}
As certain concepts are initially more entangled than others, different \glspl{cav} undergo different directional changes in the latent space. 
In order to get an insight into how individual directional changes happen, and how these changes affect the performance of individual concepts, we provide an in-depth analysis of the orthogonality and \gls{auroc} metrics of individual concepts during \gls{cav} optimization. 
Here, we fine-tune \glspl{cav} trained on the VGG16 model for the CelebA dataset. 

Fig.~\ref{fig:individual_metrics} illustrates the changes in \gls{auroc} and orthogonality of each \gls{cav} during optimization. Out of all concepts present in the CelebA dataset, we highlight three, namely the concept with the highest increase (``Pale Skin''), highest decrease (``Young''), and least change in \gls{auroc} (``Black Hair''). 
We observe that while all concepts increase in orthogonality during optimization, some experience an improvement, while others suffer from a decline in their classification performance. Furthermore, although some concepts maintain a relatively stable \gls{auroc} performance, e.g. concept with blue line on Fig.~\ref{fig:individual_metrics}, they still experience an increase in orthogonality, either due to the fact that they change their direction without losing their classification performance, or that other entangled concepts move away from the concept, such that it becomes more orthogonal to them. 

\begin{figure*}[t]
    \centering
    \includegraphics[width=1\textwidth]{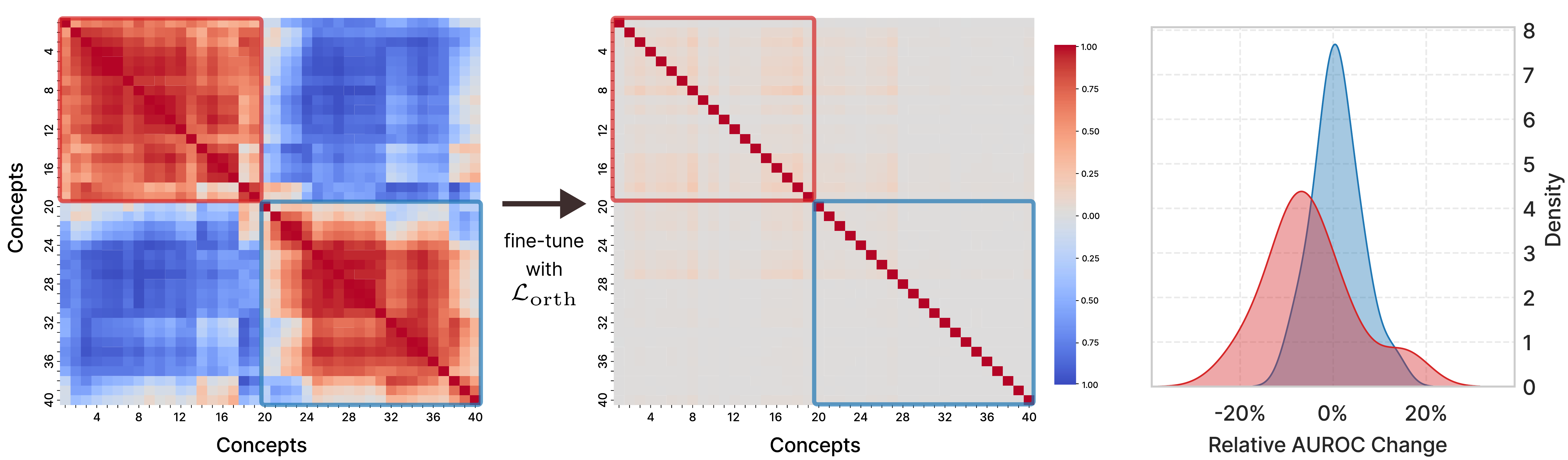}
    \caption{\emph{Left and Middle}: Cosine similarity matrices of \glspl{cav}, trained on VGG16 model on CelebA dataset, before and after the fine-tuning. \emph{Right}: Kernel Density Estimation of relative \gls{auroc} changes of individual concepts in two entangled blocks of female- (\emph{red}) and male-associated (\emph{blue}) concepts.}
    \label{fig:cos_sim_before_after}
\end{figure*}

The dynamics of entangled concepts during optimization is further highlighted in Fig.~\ref{fig:cos_sim_before_after}, where we show the \gls{auroc} change distribution of two different sets of entangled concepts before and after the optimization on the CelebA dataset starting from an optimal state for \glspl{cav}. The sets are chosen from the initial cosine similarity matrix of \glspl{cav} that are optimized without any orthogonalization, where we observe a naturally emergent blocks due to correlations in dataset associated with female- and male-related concepts. 
The resulting \gls{auroc} change distribution shows that in each entanglement block, there are concepts that experience an increase in \gls{auroc}, concepts that suffer from a decrease in \gls{auroc}, and concepts that are relatively stable in terms of \gls{auroc}, which is a clear indication that \glspl{cav} undergo a redistribution of representational importance. 
Some concepts sacrifice their distinctiveness or classification ability to reduce redundancy and entanglement, while others gain clarity and improved directional correctness as a result of orthogonalization. 
Fig.~\ref{fig:cos_sim_before_after} further depicts a global image of concept orthogonalization in latent space, where we consider the cosine similarity matrices of \glspl{cav} before and after the optimization. 
The resulting cosine similarity matrix after orthogonalization illustrates the achievement of the secondary goal of optimization, \ie, concept disentanglement, as it resembles the identity matrix, which was the desired root point of $\lorth$ defined in Section~\ref{non_orthogonality_penalization}.
Similar results are obtained for the FunnyBirds dataset, as shown in Appendix~\ref{appendix:funnybirds_details}.

\begin{figure*}[t]
    \centering
    \includegraphics[width=0.65\textwidth]{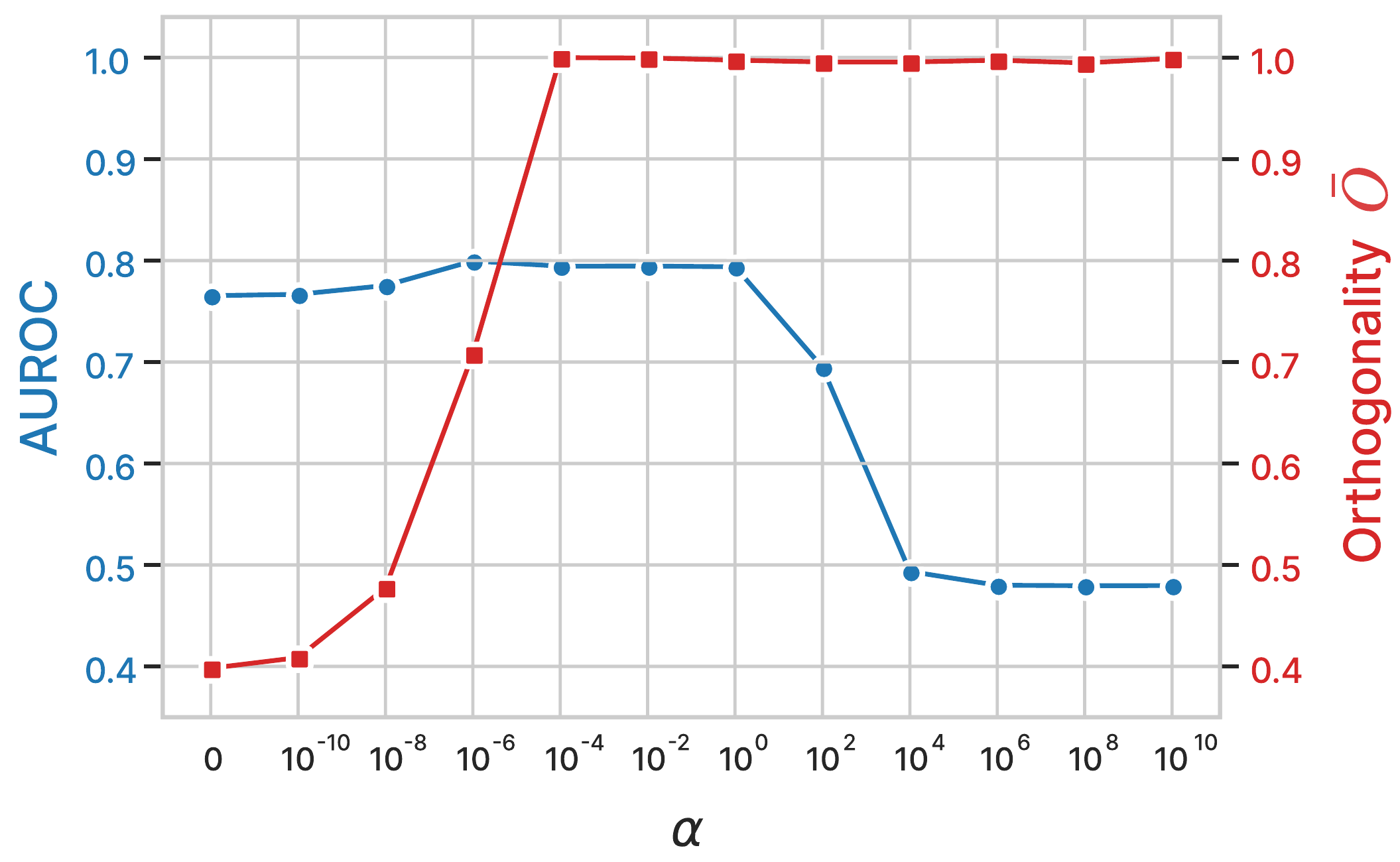}
    \caption{The average orthogonality $\bar{O}$ and \gls{auroc} after \gls{cav} training on the last convolutional layer of VGG16 for CelebA. 
    The x-axis represents the magnitude of $\alpha$, while AUROC (blue) and orthogonality (red) are shown on the y1- and y2-axis, respectively.}
    \label{fig:different_alphas}
\end{figure*}

\paragraph{Impact of Weighting Parameter $\alpha$.}
We measure the orthogonality and \gls{auroc} of \glspl{cav} after optimization with our disentanglement loss introduced in Eq.~\eqref{eq:total_loss} with various weightings \mbox{$\alpha \in [10^{-10}, 10^{-9}, ..., 10^{10}]$}. 
We run the optimization for randomly initialized \glspl{cav} for the VGG16 model for CelebA with a learning rate of $0.1$ and $500$ epochs. 
The results are shown in Fig.~\ref{fig:different_alphas}, where the AUROC and orthogonality metrics are displayed for each value of the weighting parameter $\alpha$. 
The results indicate that higher $\alpha$ values expectedly lead to a better level of orthogonalization and plateau when perfect orthogonalization is achieved. 
Interestingly, we observe that for small weighting parameters, the average AUROC slightly \emph{increases} compared to the baseline \glspl{cav} without orthogonality constraint ($\alpha=0$).
This reflects the findings discussed in Section~\ref{orthogonalization_directional_correctness}, \ie, the joint optimization of \emph{all} concept directions simultaneously can positively impact the directional correctness.
However, with larger $\alpha$ values we observe a drastic drop in \gls{auroc}, indicating that the orthogonalization constraint outweighs the directional correctness objective. 
These results demonstrate the role of $\alpha$ to balance the trade-off between directional correctness, measured via \gls{auroc}, and the level of orthogonalization.
Our results further indicate that optimal $\alpha$ values, where both \gls{auroc} and $\bar{O}$ are maximized (here: values between $10^{-4}$ and $10^{0}$), can (1) achieve perfect concept orthogonalization, and (2) refine directional correctness and achieve higher \gls{auroc} scores. 

\subsection{Disentanglement of Concept Heatmaps}
\label{sec:concept_heatmaps}
\begin{figure*}[t]
    \centering
    \includegraphics[width=.9\textwidth]{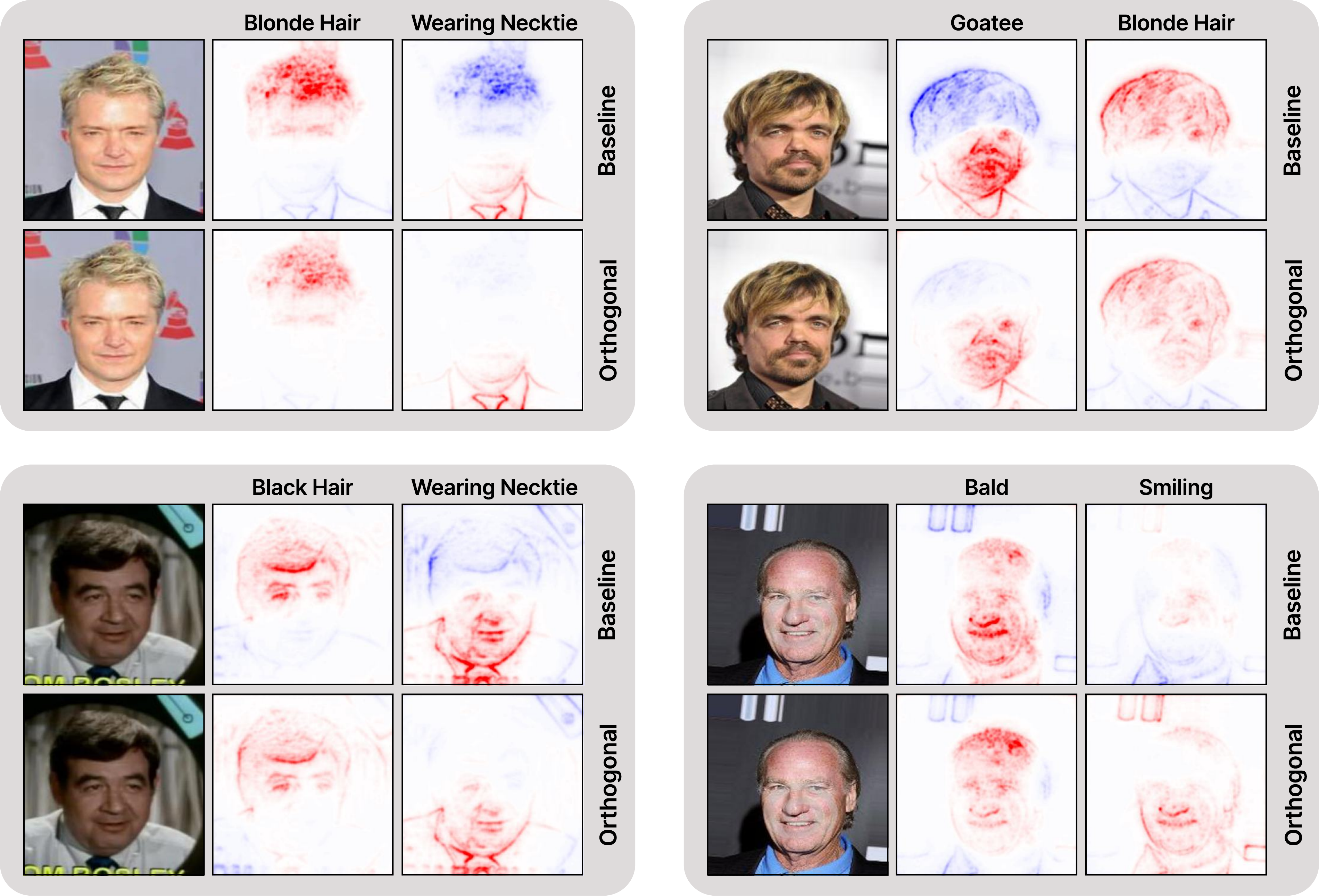}
    \caption{Concept heatmaps for 4 entangled pairs of concepts obtained using \gls{lrp} on the VGG16 model for CelebA. 
    For each pair, the three columns represent the original image (\emph{left}) and the heatmaps corresponding to first (\emph{middle}) and second (\emph{right}) concept, respectively, whereas the rows represent the heatmaps obtained before and after \gls{cav} orthogonalization. 
    Red and blue regions indicate positive and negative relevance.}
    \label{fig:heatmaps}
\end{figure*}

Although evaluating concept disentanglement through orthogonality and \gls{auroc} metrics offer valuable insight into the quantitative performance of our method, they come short in providing a qualitative understanding of the underlying representational changes of concepts. To address this, we utilize \gls{lrp} \cite{bach2015pixel} to generate heatmaps for entangled concept pairs before and after orthogonalizing the \glspl{cav} using the \texttt{zennit} library \cite{anders2021software}. 
More specifically, we compute the inner product between the activation of a sample and the \gls{cav} of interest, and then backpropagate the resulting relevances back to the input space  
\cite{pahde2023reveal,anders2022finding,pahde2025ensuring}. Similarly, other attribution methods such as Guided Backpropagation \cite{springenberg2014striving}, SmoothGrad \cite{smilkov2017smoothgrad}, and Integrated Gradients \cite{sundararajan2017axiomatic} can be used to obtain such local explanations \cite{brocki2019concept}. 
The heatmaps obtained from the VGG16 model, trained on the CelebA dataset, are illustrated in Fig.~\ref{fig:heatmaps}. The baseline \glspl{cav} successfully identify the regions associated with each concept; however, they show limitations in isolating the concepts entirely. 
Specifically, due to inherent correlations in the dataset, such as the high negative correlation observed between ``blond hair'' and ``necktie'' attributes, the heatmaps display negative relevance in each others regions that are unrelated to the original concept. 
In other words, the model tends to rely on the presence of one concept to indicate the absence of the other, or vice versa. 
Conversely, following the orthogonalization of the \glspl{cav}, the heatmaps continue to accurately highlight the relevant regions, thereby demonstrating directional correctness, while substantially reducing the negative correlations and improving the isolation of individual concepts. This outcome indicates that \glspl{cav} obtained by our method not only effectively capture the relevant spatial regions within the input space but also achieve superior performance in disentangling and isolating the concepts.

\section{CAV-based Activation Steering Applications}
\label{applications}
To evaluate the benefits of orthogonalized \glspl{cav}, we compare the results for CAV-based steering tasks using baseline \glspl{cav} without non-orthogonality penalization and orthogonalized \glspl{cav}. 
Specifically, we use \glspl{cav} to model concept directions in the latent space  to either  \emph{insert} isolated concepts into input samples using generative models in qualitative experiments (Section~\ref{concept_addition_generative_models}), or to \emph{remove} targeted concepts during inference time for shortcut suppression in a controlled scenario with both quantitative and qualitative results (Section~\ref{shortcut_removal}). 

\begin{figure*}[t]
    \centering
    \includegraphics[width=\textwidth]{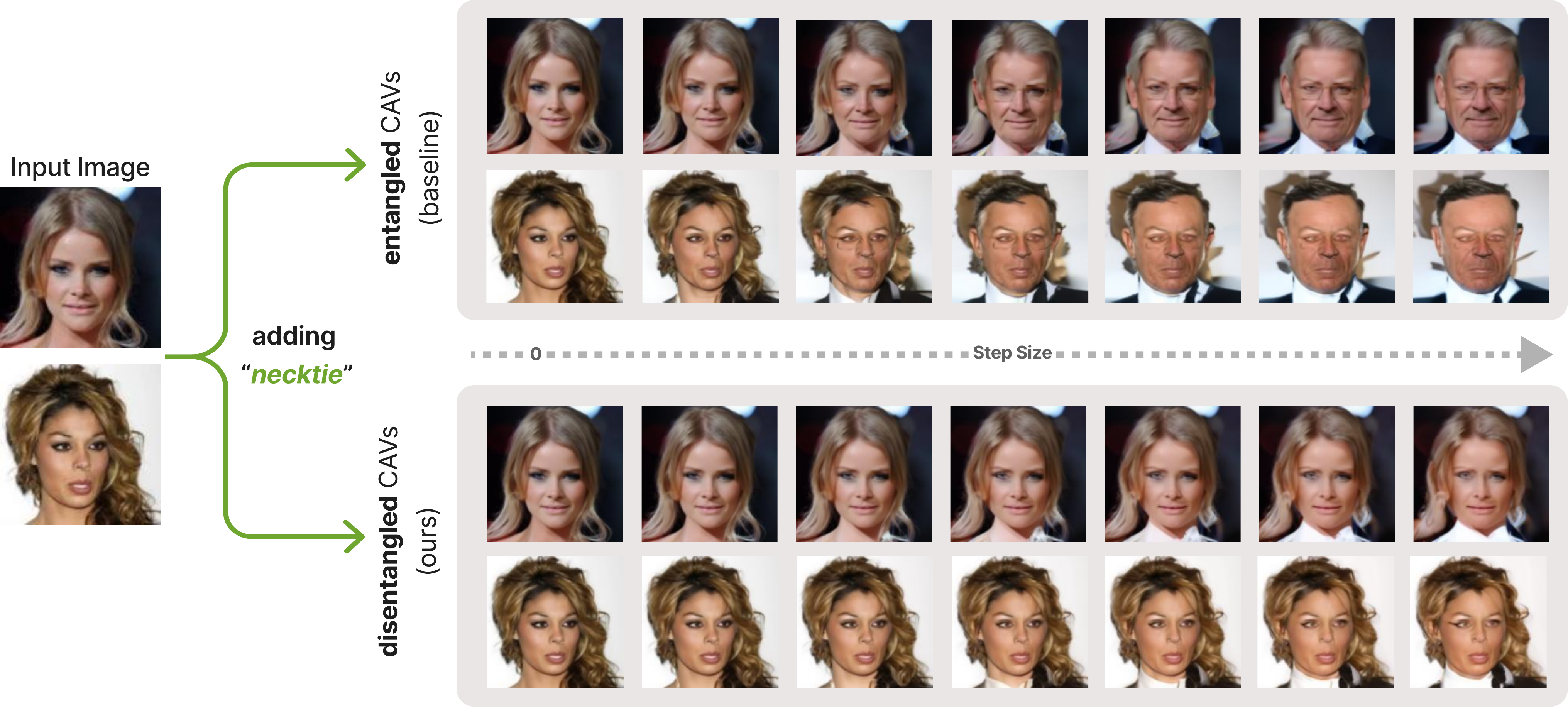}    \caption{
    Given input images (\emph{left}), we utilize a Diffusion Autoencoder to reconstruct manipulated latent encodings obtained at the bottleneck layer of the Diffusion Autoencoder. 
    Specifically, we add activations with different step sizes into the concept direction for ``necktie'', as modeled via baseline (\emph{top}) and orthogonalized (\emph{bottom}) \glspl{cav}.
    Whereas entangled baseline \glspl{cav} add correlated concepts in addition to the target concept, the disentangled \gls{cav} is capable of adding the ``necktie'' in isolation with minimized impact on other concepts.}
    \label{fig:diffae_results}
\end{figure*}
\subsection{Concept Insertion with Generative Models}
\label{concept_addition_generative_models}
As a first steering task, we consider the insertion of concepts by adding activations into the \gls{cav} direction.
Intuitively, if precisely modeled, this corresponds to adding the concept to the input image.
However, if concept representations are entangled, such as ``wearing necktie'' and ``mustache'' concepts in CelebA, adding activations along the \gls{cav} directions might lead to undesired effects.
To investigate this qualitatively, we utilize a Diffusion Autoencoder~\cite{preechakul2022diffusion} trained on CelebA dataset. Diffusion Autoencoders are a class of generative models that learn structured latent representations by combining a learnable encoder with a diffusion-based decoder. Unlike traditional autoencoders, Diffusion Autoencoders separate the latent space into two components: a semantic subcode, which encodes high-level, structured information, and a stochastic subcode, which models low-level stochastic variations. This separation allows Diffusion Autoencoders to achieve both meaningful representation learning and high-level reconstruction capabilities, making them well-suited for activation steering applications. 
This allows the generation of images representing the manipulated encoding with added activations, \ie, inserted concepts.
 
We train both \emph{baseline} and \emph{orthogonalized} \glspl{cav} on the bottleneck layer of the Diffusion Autoencoder, using the pretrained model weights provided by the authors of~\cite{preechakul2022diffusion}.
Subsequently, we utilize the learned \gls{cav} for the concept ``wearing necktie'' to insert the target concept by adding activations along its direction with different step sizes. Lastly, we decode the manipulated encoding to obtain the corresponding images. 

Fig.~\ref{fig:diffae_results} illustrates the resulting images for both \gls{cav} directions over different step sizes. 
Both \glspl{cav} succeed in adding the desired concept to given samples.
However, the baseline \gls{cav} erroneously adds other \emph{entangled} concepts, such as ``mustache'' and ``eyeglasses'', as a result of correlated concept directions. 
In contrast, the orthogonalized \gls{cav} shows superior results by adding the desired concept whilst avoiding the addition of correlated concepts.
In addition, the results indicate that our proposed approach generalizes well to larger models, such as Diffusion Autoencoders, demonstrating its scalability and effectiveness in more complex architectures.

\begin{figure*}[t]
    \centering
    \includegraphics[width=\textwidth]{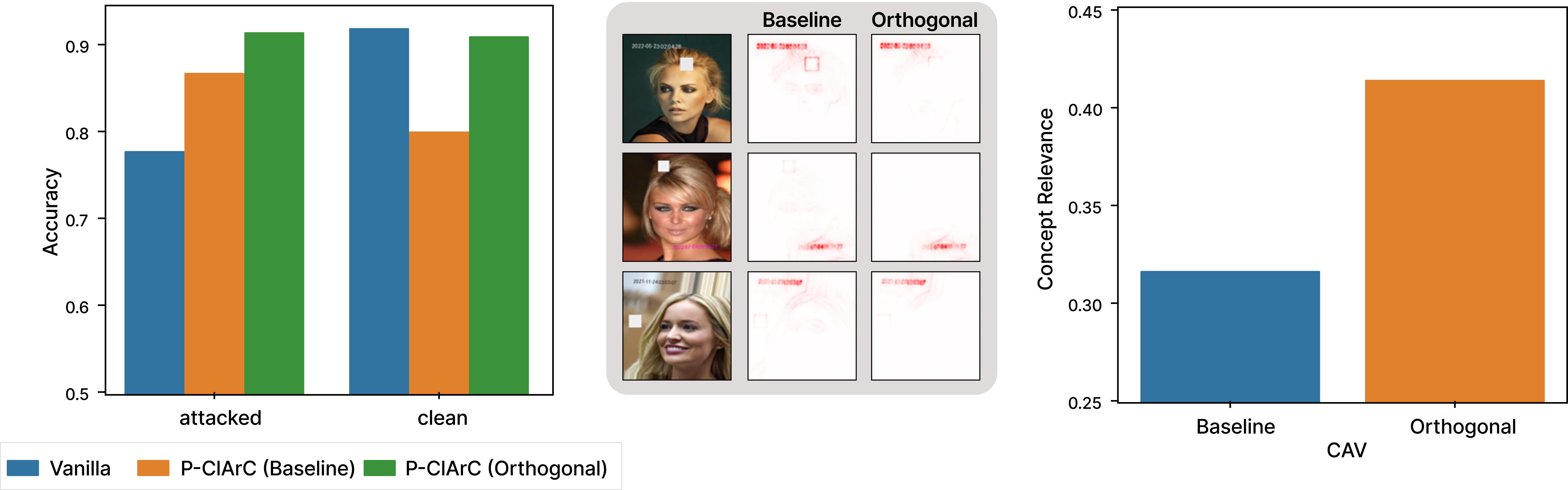}    \caption{
    \emph{Left}: Accuracy on clean and attacked test sets for the Vanilla model and \gls{pclarc}-corrected models using baseline- and orthogonal-\glspl{cav}. The latter achieves a higher robustness towards the artifact and less collateral damage. 
    \emph{Middle}: Concept heatmaps for the timestamp.
    \emph{Right}: Percentage of relevance for concept heatmaps on ground truth region. The orthogonal CAV localizes the concept more precisely.
    }
    \label{fig:pclarc_results}
\end{figure*}
\subsection{Quantifying Collateral Damage: Concept Removal for Shortcut Suppression}
\label{shortcut_removal}
Another popular steering task with \glspl{cav} is the removal of undesired concepts for the suppression of shortcut behavior. 
Shortcuts can be caused by spurious correlations in the training data, where concepts (causally) unrelated to the task correlate with the target label~\cite{geirhos2020shortcut}. 
Post-hoc model editing approaches for shortcut suppression utilize concept representations, such as \glspl{cav}, to model the undesired data artifact in latent space and suppress activations related to the concept during inference~\cite{ravfogel2020null,anders2022finding,neuhaus2023spurious,bareeva2024reactive}.
However, model editing with entangled concept representations can lead to collateral damage, as concepts entangled with the targeted concept are harmfully suppressed as well~\cite{kumar2022probing,bareeva2024reactive}. 
In this experiment, we measure both the effectiveness of the shortcut suppression and the caused collateral damage for the model editing approach \gls{pclarc}~\cite{anders2022finding} which subtracts activations in the \gls{cav} direction at inference time.
We  compare the results for the baseline and orthogonalized \glspl{cav} in a controlled setting using artificial artifacts in CelebA.
Specifically, we insert a ``timestamp'' as text overlay into $40\%$ of samples of class ``blonde'' and only $0.5\%$ of samples of class ``non-blonde'', causing a spurious correlation between the ``timestamp'' and the class label. 
In addition, we insert a ``white box'' with varying size at random positions with probability $50\%$ if there is a ``timestamp'' and $10\%$ otherwise, leading to an entanglement between the concepts ``timestamp'' and ``box''.
We train a VGG16 model with stochastic gradient descent and a learning rate of $0.001$, and expect the model to learn a shortcut based on the artificial correlation.
To measure the model's sensitivity towards the concept, we evaluate the performance on an attacked test set with the timestamp inserted into \emph{all} samples. 
This distribution shift causes models utilizing the shortcut to perform poorly.
Further, using a clean test set with \emph{no} artificial artifacts, allows evaluating the caused collateral damage.
We compute baseline \glspl{cav} for the original 40 concepts, as well as the newly introduced concepts ``box'' and \mbox{``timestamp''}, using the Pattern-CAV defined in Eq.~\eqref{eq:signal-cav}. 
For the concept disentanglement, we fine-tune the pre-trained \glspl{cav} with the orthogonalization loss in Eq.~\eqref{eq:alpha_beta_loss} with $\alpha=1$, $\beta=100$, and learning rate $0.0001$ for $500$ epochs.

The resulting classification performance on both test sets using the original model without shortcut suppression (\emph{Vanilla}) and the models corrected via \gls{pclarc} using \emph{baseline} and \emph{orthogonal} \glspl{cav} are shown in Fig~\ref{fig:pclarc_results} (\emph{left}).
The performance on the attacked test set is poor compared to the clean test set for the Vanilla model, confirming that the model uses the expected shortcut. 
Whereas the \gls{pclarc}-corrected model using the baseline \gls{cav} performs slightly better on the attacked test set, the accuracy drops significantly on the clean test set due to collateral damage. 
In contrast, the model corrected via \gls{pclarc} with the orthogonal \gls{cav} performs well on both test sets.
In addition, Fig.~\ref{fig:pclarc_results} (\emph{middle}) presents concept heatmaps using the timestamp \glspl{cav} as described in Sec.~\ref{sec:concept_heatmaps}.
Whereas the heatmap for the baseline \gls{cav} highlights the timestamp as well as correlated concepts (\eg, the ``box''), the orthogonal \gls{cav} focuses on the timestamp in isolation.
Lastly, quantifying these findings,  Fig.~\ref{fig:pclarc_results} (\emph{right}) shows the percentage of relevance on the ground truth region of the timestamp, confirming the visual findings that heatmaps computed using the orthogonal \gls{cav} provide a more accurate localization of the concept.

\section{Conclusions and Future Work}
\label{discussion}
In this work, we introduce a post-hoc concept disentanglement approach that utilizes a novel \gls{cav} training objective penalizing non-orthogonality while preserving directional correctness. 
We show the applicability of our loss term to find \glspl{cav} that are both meaningful and orthogonal with controlled and real-world concept correlations.
We further demonstrate the benefits of orthogonalized concept directions for \gls{cav}-based steering tasks, allowing to add or remove concepts in isolation without impacting correlated concepts.
Future research might focus on the semi-supervised training of \glspl{cav} to identify unlabeled concepts in addition or the definition of new training objectives allowing the joint training of multiple \gls{cav} directions.
Moreover, the integration of our orthogonalization loss into the model training is a promising research direction.

\section*{Acknowledgements}
This work was supported by
the Federal Ministry of Education and Research (BMBF) as grant BIFOLD (01IS18025A, 01IS180371I);
the European Union’s Horizon Europe research and innovation programme (EU Horizon Europe) as grants [ACHILLES (101189689), TEMA (101093003)];
and the German Research Foundation (DFG) as research unit DeSBi [KI-FOR 5363] (459422098).

\bibliographystyle{plain}
\bibliography{main}

\clearpage
\appendix
\section{CelebA Details}
\label{appendix:celeba_details}
\begin{figure*}[t]
    \centering
    \includegraphics[width=1\textwidth]{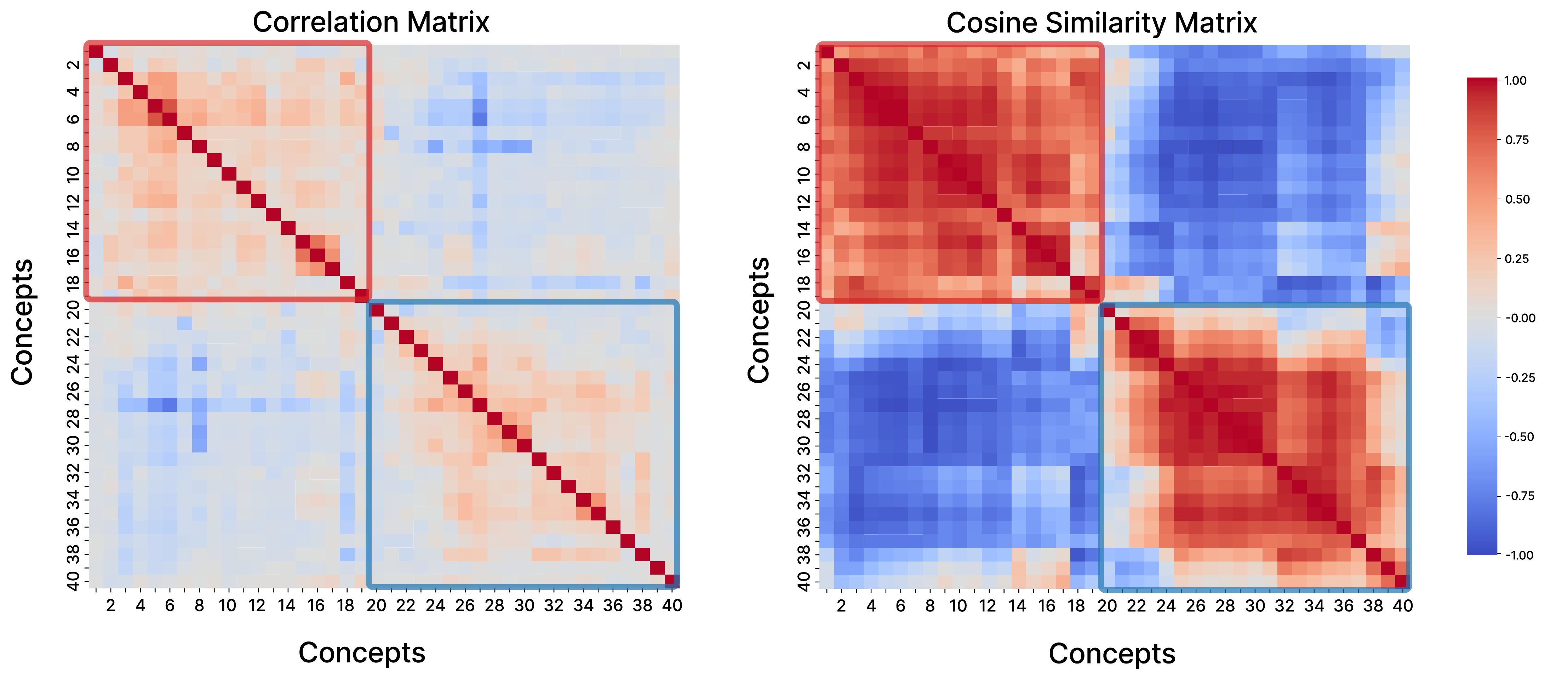}
    \caption{\emph{Left}: Correlation matrix of all concepts in CelebA. \emph{Right}: Cosine similarity matrix of concepts fitted on the last convolution layer of VGG16 trained on CelebA dataset. The two highlighted blocks on both matrices indicate the two natural groups of female (\emph{red}) and male (\emph{blue}) associated concepts.}
    \label{fig:cos_sim_vs_correaltion}
\end{figure*}

In this section we provide additional details on the CelebA dataset. 
Figure~\ref{fig:cos_sim_vs_correaltion} presents the correlation and cosine similarity matrices for concepts extracted from the last convolutional layer of a VGG16 model trained on CelebA. 
The highlighted blocks in both matrices reveal two natural concept groups associated with female (\emph{red}) and male (\emph{blue}) attributes. 
The concept labels of these groups are listed in Table~\ref{tab:celeba_concept_labels} using the same color coding for both entanglement blocks. 

\section{FunnyBirds Details}
\label{appendix:funnybirds_details}
\begin{figure*}[h]
    \centering
    \includegraphics[width=1\textwidth]{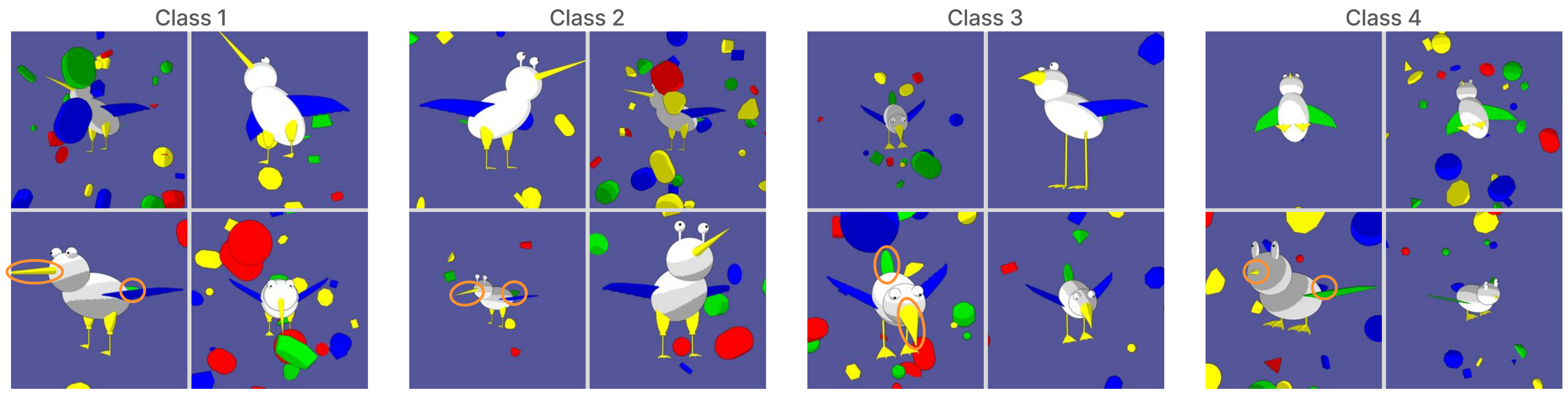}
    \caption{Example samples generated in the first $4$ classes of the FunnyBirds dataset. Some examples of concepts with enforced correlations are highlighted. }
    \label{fig:funnybirds_samples}
\end{figure*}

\begin{figure*}[h]
    \centering
    \includegraphics[width=1\textwidth]{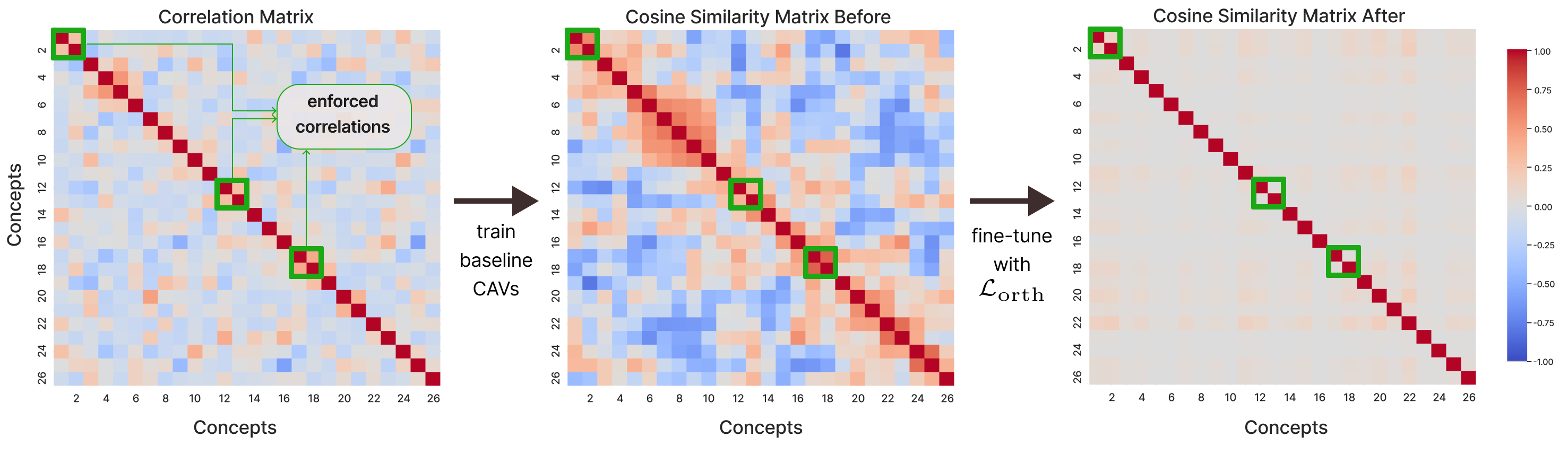}
    \caption{\emph{Left}: Correlations of known concepts based on their co-occurrence in FunnyBirds dataset. \emph{Middle and Right}: Cosine similarity matrices of \glspl{cav}, trained on ResNet18 model on the FunnyBirds dataset, before and after the fine-tuning. The highlighted blocks indicate the concept pairs with enforced correlations.}
    \label{fig:corr_cos_funnybirds}
\end{figure*}

\begin{table}[h]
    \centering
    \footnotesize
    \renewcommand{\arraystretch}{1.2}
    \begin{tabular}{|l|c|c|c|}
        \hline
        \textbf{Beak Model} & \textbf{tail01.glb} & \textbf{tail02.glb} & \textbf{tail03.glb} \\
        \hline
        beak01.glb & 0.7  & 0.15 & 0.15 \\
        beak02.glb & 0.15 & 0.7  & 0.15 \\
        beak03.glb & 0.15 & 0.15 & 0.7  \\
        beak04.glb & 0.15 & 0.15 & 0.7  \\
        \hline
    \end{tabular}
    \vspace{5pt}
    \caption{Correlations between beak models and tail models based on enforced probabilities during dataset generation.}
    \label{tab:beak_tail_correlations}
\end{table}
We provide details about the synthetic FunnyBirds dataset we generated for our experiments. In Table~\ref{tab:beak_tail_correlations} we list the probabilities we used for the co-occurrence of concepts during the dataset generation to enforce correlations in the dataset, namely between \emph{beak} and \emph{tail} concepts. The resultant correlations, as well as the cosine similarity matrices before and after the \gls{cav} fine-tuning are illustrated in Fig.~\ref{fig:corr_cos_funnybirds}, where we fine-tune \glspl{cav} on the ResNet18 model trained on the FunnyBirds dataset (see Appendix~\ref{appendix:cav_training_details} for more details).
Finally, in Fig.~\ref{fig:funnybirds_samples} we display some example samples for the first 4 classes out of 50 total classes.
\newpage
\section{\gls{cav} Training Details}
\label{appendix:cav_training_details}
\begin{table}[h]
    \centering
    \small
    \renewcommand{\arraystretch}{1.2}
    \begin{tabular}{|l l c c c|}
        \hline
        \textbf{Dataset} & \textbf{Model} & \textbf{LR} & $\boldsymbol{\alpha}$ & \textbf{Epochs} \\
        \hline
        {CelebA} & VGG16 & $0.001$ & $0.01$ & $300$ \\
                 & ResNet18 & $0.001$ & $0.01$ & $300$\\
        \hline
        {FunnyBirds} & VGG16 & $0.001$ & $0.1$ & $300$ \\
                    & ResNet18 & $0.001$ & $0.01$ & $300$ \\
        \hline
    \end{tabular}
    \vspace{5pt}
    \caption{Hyperparameters used for \gls{cav} fine-tuning. $\alpha$ denotes the weighting parameter used in Eq.~\eqref{eq:orth_cav}.}
    \label{tab:cav_hyperparams}
\end{table}
In this section, we provide the hyperparameters used for training \glspl{cav}. Table~\ref{tab:cav_hyperparams} lists the learning rate, weighting parameter $\alpha$ and number of epochs used for fine-tuning \glspl{cav} for each dataset-model combination.  

\begin{table*}[h]
    \centering
    \renewcommand{\arraystretch}{1.2}
    \begin{tabular}{|c|l|c|l|c|l|}
        \hline
        \textbf{Index} & \textbf{Concept} & \textbf{Index} & \textbf{Concept} & \textbf{Index} & \textbf{Concept} \\
        \hline
        1  & \textcolor{red}{Pale Skin} & 15  & \textcolor{red}{High Cheekbones} & 29  & \textcolor{blue}{Goatee} \\
        2  & \textcolor{red}{Oval Face} & 16  & \textcolor{red}{Smiling} & 30  & \textcolor{blue}{Sideburns} \\
        3  & \textcolor{red}{Attractive} & 17  & \textcolor{red}{Mouth Slightly Open} & 31  & \textcolor{blue}{Wearing Necktie} \\
        4  & \textcolor{red}{Arched Eyebrows} & 18  & \textcolor{red}{Young} & 32  & \textcolor{blue}{Receding Hairline} \\
        5  & \textcolor{red}{Heavy Makeup} & 19  & \textcolor{red}{Big Lips} & 33  & \textcolor{blue}{Bald} \\
        6  & \textcolor{red}{Wearing Lipstick} & 20  & \textcolor{blue}{Brown Hair} & 34  & \textcolor{blue}{Double Chin} \\
        7  & \textcolor{red}{Wavy Hair} & 21  & \textcolor{blue}{Straight Hair} & 35  & \textcolor{blue}{Chubby} \\
        8  & \textcolor{red}{No Beard} & 22  & \textcolor{blue}{Black Hair} & 36  & \textcolor{blue}{Eyeglasses} \\
        9  & \textcolor{red}{Wearing Necklace} & 23  & \textcolor{blue}{Bushy Eyebrows} & 37  & \textcolor{blue}{Wearing Hat} \\
        10  & \textcolor{red}{Pointy Nose} & 24  & \textcolor{blue}{5 o'Clock Shadow} & 38  & \textcolor{blue}{Gray Hair} \\
        11  & \textcolor{red}{Rosy Cheeks} & 25  & \textcolor{blue}{Bags Under Eyes} & 39  & \textcolor{blue}{Blurry} \\
        12  & \textcolor{red}{Wearing Earrings} & 26  & \textcolor{blue}{Big Nose} & 40  & \textcolor{blue}{Narrow Eyes} \\
        13  & \textcolor{red}{Bangs} & 27  & \textcolor{blue}{Male} &  &  \\
        14  & \textcolor{red}{Blond Hair} & 28  & \textcolor{blue}{Mustache} &  &  \\
        \hline
    \end{tabular}
    \vspace{5pt}
    \caption{Concept labels present in the CelebA dataset, ordered and colored as in the correlation and cosine similarity matrices (best seen in color).}
    \label{tab:celeba_concept_labels}
\end{table*}

\end{document}